# Detection of false positives by human reviewers depends on the face recognition algorithm and its security configuration

**Esha Edakoth, John J. Howard, Isabelle M. Shuggi*, Yevgeniy B. Sirotin, Robert Wood**

**Authors listed alphabetically.**

***Corresponding author: ishuggi@idslabs.org**

## ABSTRACT

Integration of artificial intelligence (AI) technology into identity workflows, such as automated face recognition (FR), has enhanced operational capabilities, with FR algorithms achieving super-human scale and accuracy. However, these gains can make human oversight of FR more difficult, raising questions about the effectiveness of these human-AI teams. We measured human reviewers' ability to detect false positive errors made by commercial FR algorithms across varying security configurations. Using a novel dataset, the Difficult Facial Image Comparison Test (DFICT), derived from false positives made by four comparable commercial face recognition algorithms, we measured the ability of 801human reviewers to detect algorithmic false positives across three security configurations, representing operationally relevant biometric thresholds. Our findings show that human reviewers can reduce algorithmic false positives, but performance declines at stricter security thresholds. Further, reviewer performance varied by algorithm indicating that some algorithms are more compatible with human oversight. We suggest that biometric systems should be designed to account for human reviewers' variable effectiveness, particularly in high-security applications. Our findings should be considered during algorithm selection and inform human oversight strategies to minimize AI-related risks in operational FR deployments.

## INTRODUCTION

The application of human-AI teams to optimize task performance has become common across industries. As AI technology improves, the nature of human-AI teams continues to evolve enabling richer interactions but also raising questions about the role and value of each teammate. AI has been leveraged in identity workflows to improve the ability of such systems to reliably identify individuals based on previously enrolled biometric samples. These AI systems have enabled rapid and accurate identification of individuals in government use cases, such as border control. Before the advent of deep learning methods, AI based identity systems relied on the comparison of manually engineered features and automated systems achieved high accuracy comparing images of the iris and fingerprints [1,2]. Though AI is capable of low error rates, the scale of application in some systems could still result in errors [3]. To help reduce errors, AI decisions were validated by human reviewers. Because humans do not natively have perceptual systems to compare fingerprints or iris images, human reviewers had to be specially trained on how to adjudicate AI decisions for these biometric modalities and/or how to address the broader context of the identity decision [4]. Together AI algorithms and human reviewers formed a human-AI team.

The application of deep learning methods to facial recognition has dramatically improved the performance of automated FR technology. Consequently, FR technology has been increasingly deployed in government identity use cases, which has strong implications for human-AI teaming since humans have a distinct neural system specialized for performing facial comparisons. In identity workflows, human face recognition has been traditionally leveraged to verify the identity of individuals based on a photo ID [5]. When humans team with FR technology, they are typically placed in a supervisory role where they review the matching decisions made by an algorithm, even without standardized training on facial comparison tasks. The notion that human reviewers of FR algorithms can be effective without specialized training stems from the fact that humans use facial comparison to recognize conspecifics [6]. However, recent FR false positives have raised questions about the effectiveness of human-AI teams in face recognition workflows [7–9].

**Human Face Recognition**

Prior research has shown that parts of the human brain are specialized for face recognition which give humans an innate, holistic, face recognition ability [10–12]. Humans innately have the capability to recognize faces of both familiar and unfamiliar individuals; however, recognition of unfamiliar faces is more challenging [13,14]. For most identity workflows, human reviewers perform comparisons between faces of individuals unfamiliar to them. Differences in human capabilities to compare unfamiliar faces have been found with trained facial examiners and super-recognizers who perform the task with higher accuracy [12,15].

Humans can be trained to recognize unfamiliar faces; however, research has shown that improvements in accuracy are dependent on the methods of training [16]. Although humans process faces holistically, it has been shown that instructing inexperienced reviewers to focus on specific facial features can improve their accuracy when matching unfamiliar faces [17]. Human reviewers are also capable of leveraging non-facial features [18]. Facial examiners are trained specifically to not rely on their holistic face perception and instead focus on comparison of individual facial morphological features [19]. However, many human-AI teams in biometric workflows employ staff without formal training in facial comparison and/or do not afford the time required to perform a comprehensive morphological comparison.

**Face Recognition Technologies**

FR technologies extract features from facial images to create abstract mathematical representations (templates) that are compared numerically to create scores representing the level of similarity between two templates. FR algorithms create distinct score distributions for mated comparisons between templates of the same person and non-mated comparisons between templates of different people. A match comparison decision for a face pair is made when the similarity score exceeds a pre-configured biometric threshold value. This value is typically selected based on the desired level of security or facilitation, corresponding to different rates of false positive or false negative recognition errors. Algorithms can generally be categorized as academic or commercial, where commercial algorithms function at higher levels of accuracy and scalability. In government applications, security settings for commercial algorithms are usually selected to meet use-case dependent security objectives.

Early AI approaches to face recognition used hand crafted features to measure the similarity between faces using landmarks [20,21] or by applying matrix decomposition approaches to represent faces in a linear face space [22]. However, the recognition accuracy afforded by these approaches was not suitable for government use-cases [23,24] which typically require the technology to achieve false negative rates below 5% at false positive rates of 0.1% or lower. With the application of convolutional neural networks and deep learning to face recognition, however, the performance of these systems improved such that these systems met and exceeded human levels of face recognition performance [15,25]. Modern FR technologies can maintain false negative rates below 1% at false positive rates below 0.0001% on images that are representative of government use-cases [26]; however, performance is affected by subject demographics and the quality of image capture [27–30]. The model architectures used by modern FR technologies can vary [31,32] and performance depends on the datasets used for training [33]. Consequently, different FR technologies can make different comparison decisions for the same face pairs. Because government applications typically make use of current commercial FR algorithms, it is important to study and understand how these systems make comparison decisions.

FR technologies are used globally for different use cases of verification and identification [34–38]. Verification requires a probe image to be compared to one reference image (one-to-one comparison). Passing through a daily security checkpoint, an individual's image could be captured daily (probe image) and then compared to an image that was stored at the time of enrollment (reference image) into the system. During identification, a probe is compared to a database of references (one-to-many comparison). To determine if an individual is permitted to enter an event, an image of them captured at the time of entry (probe image) can be compared to a database containing images (reference images) of individuals who have previously been identified as threats. Identification requires higher biometric thresholds than verification to afford the same level of security (i.e., the same chance of a false positive error). Depending on the use case, both processes can be completed with different levels of automation and human oversight.

**Human-AI Teaming in Face Recognition Workflows**

It is crucial to consider the human and AI's capabilities when building a human-AI team in the identity workflow to mitigate false positive errors which can have serious consequences. Both humans and AI process visible facial features through neural networks; however, the processed features and the methods of processing are different [39]. Human reviewers can consider features that an FR algorithm may ignore such as marks, scars, or tattoos. Furthermore, different FR algorithms may rely on different features. Although each teammate can provide different strengths, prior teaming research has shown that human reviewers can paradoxically decrease team performance by overriding accurate algorithm decisions and/or failing to catch errors, especially when image quality is degraded [40–44]. Recent work suggests that different FR algorithms can change the nature of the face image pairs presented for human review and, consequently, reviewer performance [45]. However, prior work has not considered human review of the biometric decisions made by different commercial FR algorithms with comparable performance or different operationally relevant security settings.

Prior research on human-AI teaming in biometrics distinguished AI and human teammate roles based on decision-making capabilities and interdependence requirements in the context of humans teaming with biometric algorithms [46]. This work defined different levels of automation (LOAs) for five subsystems (data capture, signal processing, data storage, comparison, and decision) of the standard biometric identity workflow defined in ISO/IEC 19795-1 [47]. LOAs for each subsystem can be selected based on the requirements of a specific use case leading to the formation of teams with different types of task interdependence.

Task interdependence is determined by the dependency between teammates to complete their tasks and can influence the design of human-AI teams. Teaming research has defined four types of task interdependence: reciprocal, sequential, pooled, and team [48]. The components of a task determine the type of interdependence which in turn influences how the human-AI team should be structured. Some work in FR has considered pooled interdependence where algorithms and humans make decisions independently and results are fused to improve overall performance [15,49]. The focus of our work is on sequential task interdependence, where one teammate must complete a task before the other teammate can complete their task. Consider a scenario where a human is tasked with reviewing a FR algorithm's matching decision, the human reviewer must wait for the AI to complete the comparison and make a decision before completing a review. Task interdependence affects how tasks are allocated between teammates.

Task allocation refers to selecting the role of the human and the AI teammates depending on their respective capabilities [50]. The tasks assigned to each teammate need to be appropriately defined for optimal team performance [46,50]. It is important to consider how AI teammates can affect the performance of human teammates. Humans can develop inaccurate mental models of AI based on their experience on the team, including a failure to accurately gauge the AI teammate's capabilities, which can result in decreased team performance [40,41,43,51,52]. Adjustments to the team can be made, including changes in task allocation, AI system selection, and/or training. To implement the correct adjustments, it is crucial to evaluate the human's capabilities relative to the AI's capabilities.

**Contributions**

Minimizing false positives in operational identity workflows is critical for detecting imposters and preventing the serious consequences of false identifications, especially in high-security applications like law enforcement. In operational settings, human operators review biometric outcomes from commercial algorithms, forming human-AI teams characterized by sequential task interdependence. This study addressed three key questions:

1. Are human reviewers capable of detecting FR false positives made by current commercial algorithms?
2. How do security settings affect human reviewers' ability to catch an FR false positive?
3. Do commercial algorithms differ in their suitability for human-AI teaming?

To address these questions, we selected face pairs based on similarity scores from four comparable commercial algorithms. We configured operational security settings (biometric thresholds) for each algorithm using a dataset of high-quality passport-style images from 2,091 unique individuals. These biometric thresholds were used to identify face pairs that result in a false positive recognition. These face pairs were reviewed by human volunteers without prior exposure to algorithmic decisions to obtain an unbiased assessment of reviewers' ability to detect algorithmic false positives. This study provides the following contributions:

1. A novel method (Sampling by False Positive Rate Interval, S-FPRI) for selecting face pairs to assess human-AI team performance at operationally relevant settings.
2. Advances in our understanding of human oversight in FR workflows:
    a. Human reviewers are capable of catching FR false positives made by current commercial algorithms.
    b. At stricter security settings (i.e., higher biometric thresholds), human reviewers are less likely to catch FR false positives.
    c. Reviewer performance varies by algorithm, indicating that some algorithms may be more suitable for human-AI teaming.

## METHODS

### Face Pair Selection Methodology

Face pairs were selected using a methodology we call Sampling by False Positive Rate Interval (S-FPRI). This methodology allowed us to make performance comparisons across FR algorithms despite fundamental differences in the similarity scales used by these different AI systems. For each face recognition algorithm, we computed four biometric threshold settings (security or false match rate (FMR) settings) that produced false match rates of 1e-3, 1e-4, 1e-5, and 1e-6 based on a reference face image dataset. The four commercial algorithms we used had comparable performance with false non-match rates at or below 1% for each of the evaluated FMR settings. All images were enrollment quality collected under informed consent (approved by the WGC Institutional Review Board). To select face pairs for the survey, we filtered comparisons based on the following criteria: i) we only included unique comparisons (i.e., comparing image 1 to image 2 was considered equivalent to comparing image 2 to image 1), ii) comparisons of twins were removed, and iii) if we had more than one image per subject leading to multiple comparisons of the same subjects, only the highest scoring comparison was retained. Across the four algorithms, this filtered dataset included 2,290 images of 2,087 subjects resulting in 8,973,535 comparisons.

Non-mated pairs were stratified into three intervals based on where their similarity scores fell relative to the four FMR settings (Figure 1). The three intervals were ordered by security as follows: lower (FMR: 1e-3 to 1e-4), medium (FMR: 1e-4 to 1e-5), and higher (FMR: 1e-5 to 1e-6). We hypothesized that these False Positive Rate Intervals (FPRIs) would modulate the nominal task difficulty for human reviewers such that reviewers would be more likely to detect false positives from the lower FPRI relative to those in the higher FPRI.

Within the lower, medium, and higher FPRI, the three highest scoring non-mated pairs were selected for each algorithm and then manually reviewed to confirm that that same identity did not appear more than once within the selected pairs for that algorithm. For one of the four algorithms, we did not use the highest scoring non-mated pair in the low FPRI because one of the individuals in that pair also appeared in the highest scoring non-mated pair in the medium FPRI. We chose to replace the pair in the low FPRI with the next highest scoring non-mated pair in that bin to maintain the highest possible difficulty for the medium FPRI. We selected the three highest scoring non-mated pairs within each interval to consistently use the most challenging face pairs for humans to review that still fell within the FPRI. Furthermore, assessing human performance using non-mated face pairs close to the upper limit of each FPRI (i.e., the next highest threshold), allowed us to approximate the lower bound of human performance.

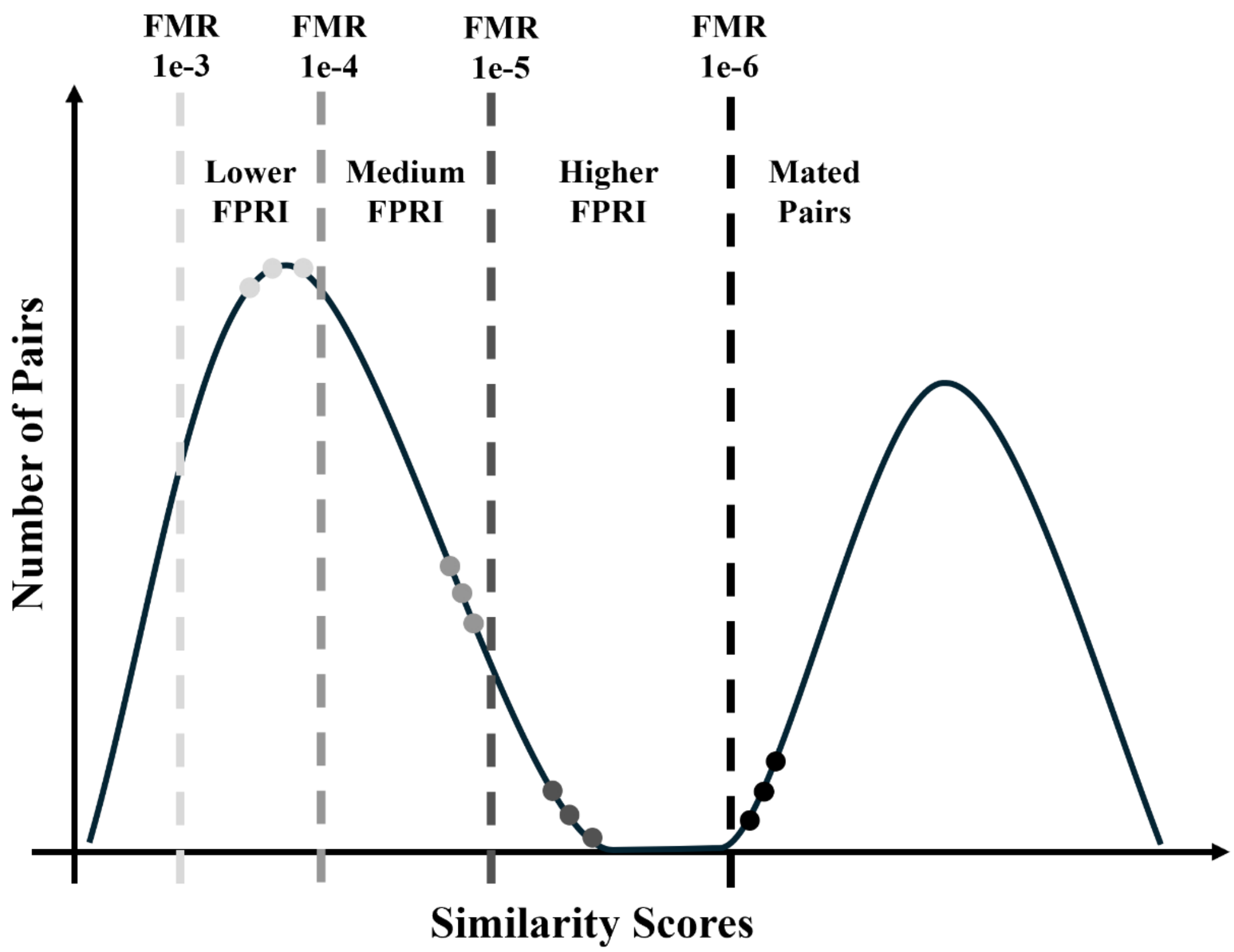

*Figure 1: Method of pair selection applied to each commercial algorithm. The dots on the similarity score distribution indicate where pairs were selected from each FPRI and each FPRI is labelled: lower (light gray dots), medium (gray dots), and higher (dark gray dots). Mated pairs are shown with black dots and were directly above the 1e-6 FMR setting.*

For each algorithm we selected the three lowest scoring mated pairs with similarity scores above the higher FPRI. Mated pairs were added to the survey to avoid habituation to the non-mated pairs and aid in the analysis of human performance. Pairs selected using the S-FPRI process were added to individual surveys of 12 questions per algorithm, resulting in four distinct surveys. Across all four surveys, there were images of 70 different individuals and 39 unique face pairs (Figure 2). One mated face pair appeared in all four of the surveys, one mated face pair appeared in three of four surveys, one of the mated pairs appeared in two of four surveys, and three non-mated pairs from the high FPRI appeared in two surveys.

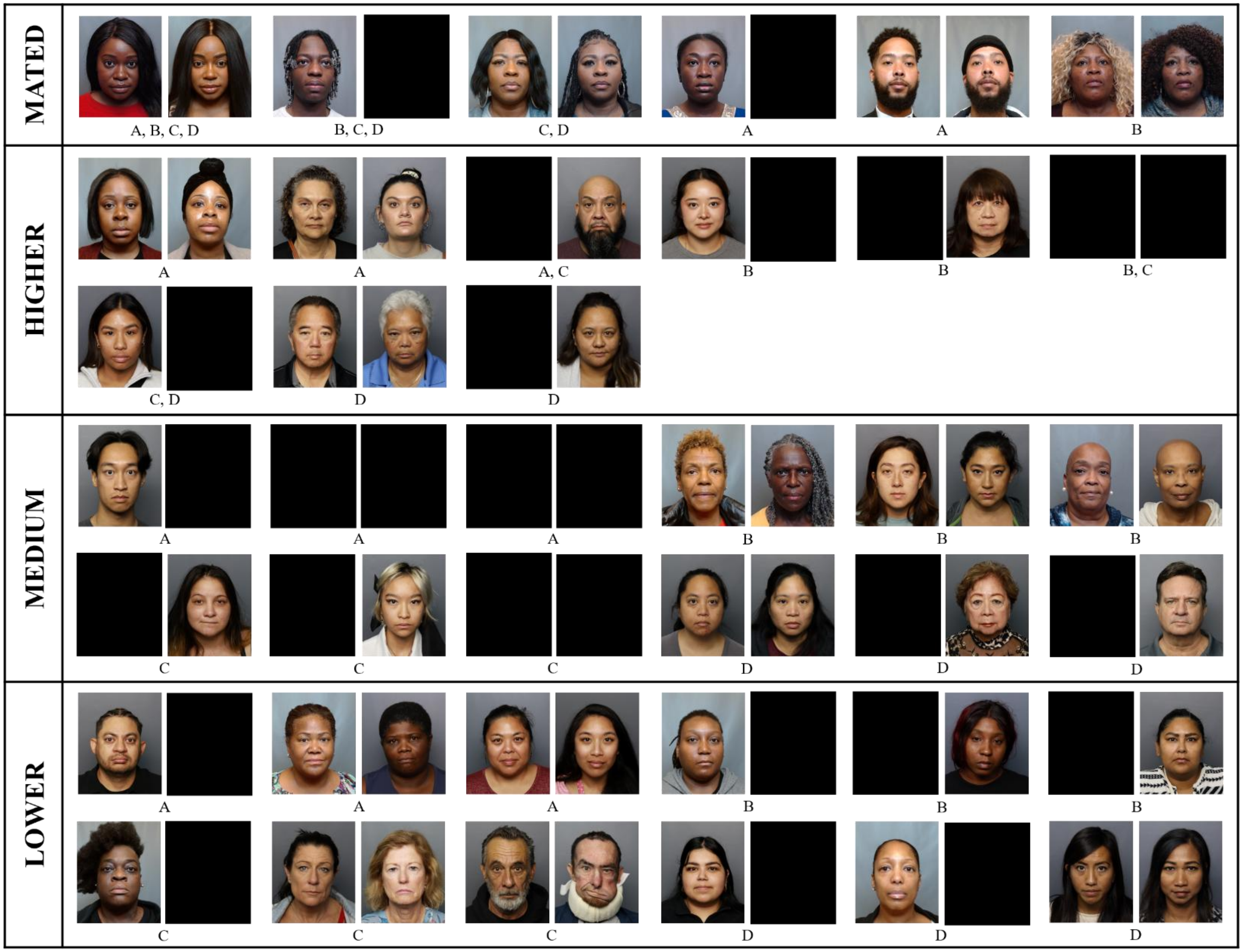

*Figure 2: Pairs from all algorithms separated by FPRI and labelled by algorithm (A, B, C, D). Each human reviewer saw 12 of these face pairs from the same algorithm. Black boxes indicate images of individuals who did not consent to images being used in publications.*

**Survey Design**

The face matching task was administered using a custom application running on Apple iPads (A1893). Volunteers were provided with task instructions but were not provided with any face recognition training. Each untrained human reviewer rated a total of 12 face pairs from one algorithm: three non-mated pairs from each FPRI and three mated pairs. Images for mated pairs were captured on different days. Every pair was selected using the S-FPRI method described above and there was one survey variant for each of the four face recognition algorithms in our study. Volunteers did not have a time limit, but most completed the entire task in a 10-minute period.

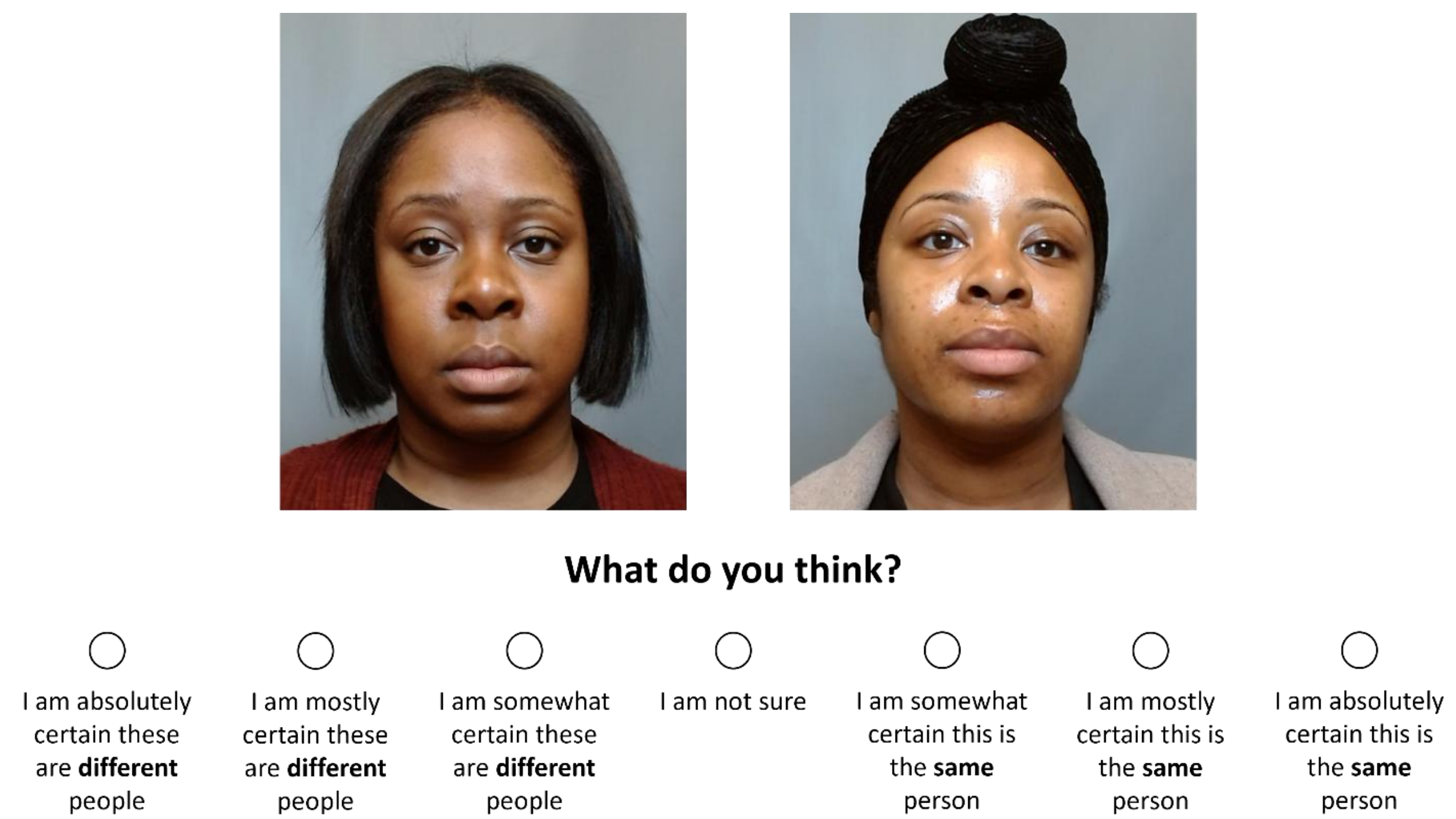

*Figure 3: Example survey question of a non-mated face pair. Subjects provided permission to share images.*

The pairs were presented one at a time with a 7-point scale ranging from “I am absolutely certain these are different people” to “I am absolutely certain this is the same person” (see Figure 3). Human reviewers were forced to select a response, with the option to select “I am not sure.” Pairs were presented in a random order regardless of their FPRI classification. A mated pair randomly appeared once every four questions.

**Volunteer Demographics**

All study materials and protocols were approved by the WGC Institutional Review Board. A total of 814 volunteers consented to participate in this study as part of two larger biometric data collection events. 13 volunteers were removed due to indications of misunderstanding the survey and/or technical difficulties with the iPads used to administer the survey. Volunteers self-reported age, sex and race to confirm balanced demographics.

*Table 1: Demographics of human reviewers*

| Demographic Category | Demographic Group | Count |
|---|---|---|
| Age Group | [18,35) | 201 |
| | [35,50) | 266 |
| | [50,65) | 202 |
| | [65,90) | 132 |
| Sex | Female | 424 |
| | Male | 374 |
| | Prefer Not to Disclose | 3 |
| Race | Black | 349 |
| | White | 250 |
| | Other | 201 |
| | Prefer Not to Disclose | 1 |

**Data Analysis**

We applied signal detection theory (SDT) to face matching tasks, which allows for an analysis of human performance that can be aligned with algorithm performance [40,43,53,54]. The two key metrics of SDT are sensitivity (*d'*) and criterion (*c*), which assess discriminability between mated and non-mated face pairs and cognitive response bias, respectively.

*Table 2: Reviewer decision thresholds.*

| θ | Description | Interpretation |
|---|---|---|
| 2.5 | Very Strict | Almost all of the pairs are different |
| 1.5 | Mostly Strict | Most of the pairs are different |
| 0.5 | Slightly Strict | Some of the pairs are different |
| -0.5 | Slightly Permissive | Some of the pairs are the same |
| -1.5 | Mostly Permissive | Most of the pairs are the same |
| -2.5 | Very Permissive | Almost all of the pairs are the same |

Application of SDT allows for the estimation of a receiver operating characteristic (ROC) curve which measures performance at different reviewer decision thresholds. The rating scale presented to human reviewers was aligned with values ranging from -3 (I am absolutely certain these are different people) to 3 (I am absolutely certain this is the same person). To compute SDT metrics, we converted certainty responses to binary decisions (1 = same, 0 = different) such that responses above each reviewer decision threshold corresponded to "same" decision and those below the decision threshold were classified as "different" decisions.

At each of the six reviewer decision thresholds defined in Table 2, we calculated the true positive rate ($TPR_\theta$) and the false positive rate ($FPR_\theta$) based on the response ($R_i$) to each question. Note that $FPR_\theta$ was calculated for each FPRI.

$$TPR_\theta = \frac{1}{m} \sum_{Mated} R_i > \theta$$

$$FPR_\theta = \frac{1}{n} \sum_{Non-mated} R_i > \theta$$

$TPR_\theta$ and $FPR_\theta$ were used to calculate the sensitivity (*d'*) and criterion (*c*) at each decision threshold.

$$d'_\theta = \Phi^{-1}[TPR_\theta] - \Phi^{-1}[FPR_\theta]$$

$$c_\theta = -\frac{\Phi^{-1}[TPR_\theta] + \Phi^{-1}[FPR_\theta]}{2}$$

The area under the ROC curve (AUC) can also be calculated as a measure of sensitivity that is independent of decision thresholds. The AUC for each FPRI was calculated based on the Gaussian signal and noise distributions assumptions [55]. In the formula below, *I* and *S* represent the intercept and slope of the best fit line between the z scores of the true positive rates ($\Phi^{-1}[TPR_\theta]$) and false positive rates ($\Phi^{-1}[FPR_\theta]$).

$$A_z = \Phi\left[\frac{I}{\sqrt{1 + S^2}}\right]$$

A supplementary metric of AUC is the calculation of *d'* across all decision thresholds.

$$d' = \sqrt{2}(\Phi^{-1}[A_z])$$

## RESULTS

### Aggregated Across Algorithms

The distributions of reviewer responses to non-mated face pairs suggest that detection of algorithmic false positives becomes progressively harder with increasing biometric thresholds from lower, to medium, and to higher FPRIs. Non-mated pair response distributions were right skewed along the perceived similarity scale. Comparing distributions across FPRIs showed that human reviewers classified face pairs as "different people" with the highest certainty most often when presented with non-mated pairs from the lower FPRI. Responses shifted to the right when reviewers were presented with non-mated pairs from the medium and high FPRIs, such that the number of responses indicating that the images were of the same person increased and the number of responses indicating high certainty that the images were of different people decreased (Figure 4).

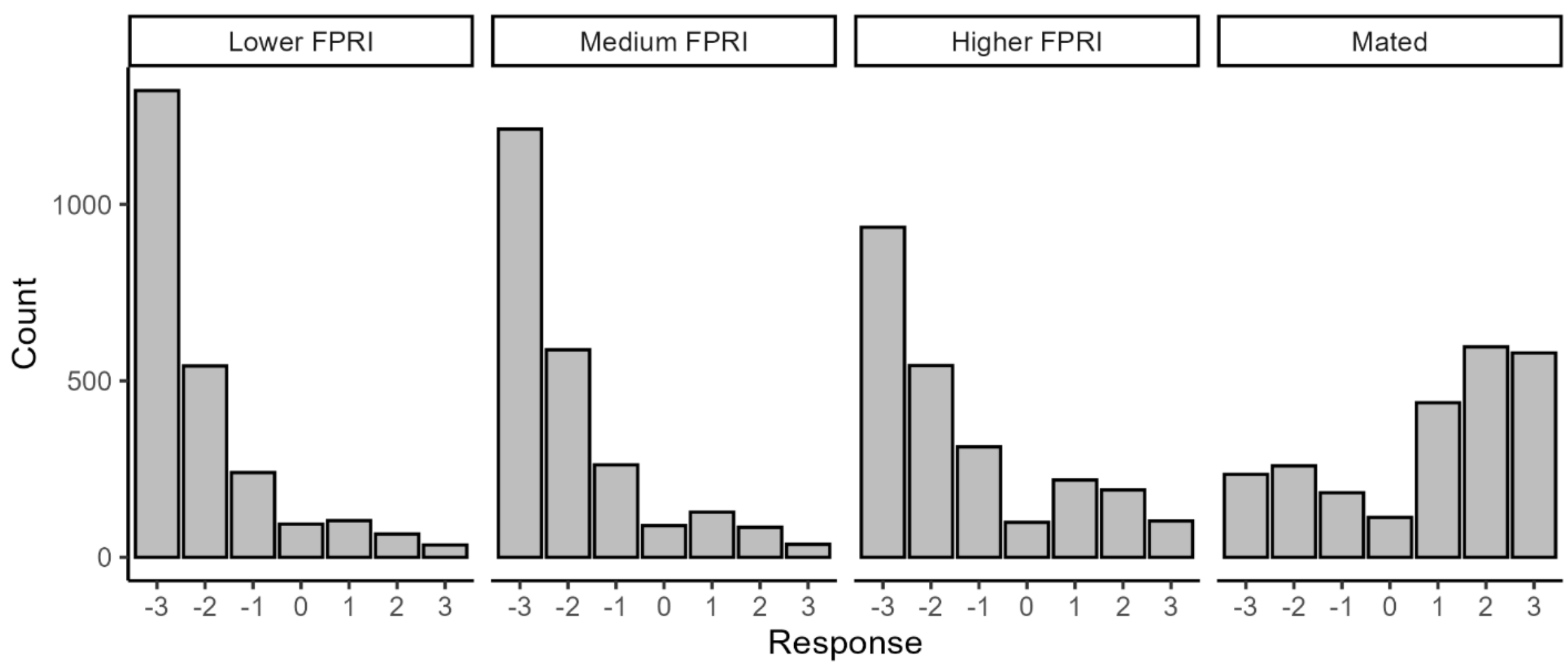

*Figure 4: Distributions of responses for each FPRI and mated pairs. Negative values indicate that the reviewer thought the images were of two different people, positive values indicate that the reviewer thought the images were of the same person and zero indicates that the reviewer was not sure. Certainty of the response is indicated by the absolute value of the response (e.g., -3 indicates absolute certainty that the images are of different people; 3 indicates absolute certainty that the images are of the same person).*

When presented with mated pairs, most human reviewers correctly classified the images as being of the "same person" but with varying degrees of certainty. Responses to mated pairs were left skewed along the perceived similarity scale. Notably, the peak of the response distribution was much lower and flatter compared to the peaks of the non-mated response distributions. These results were further assessed through the calculation of TPR and FPR for human reviewer decisions. All results are reported for the slightly strict reviewer decision threshold (θ = 0.5), separating responses that label pairs as the same or different and classifying responses of "I am not sure" as a response of different.

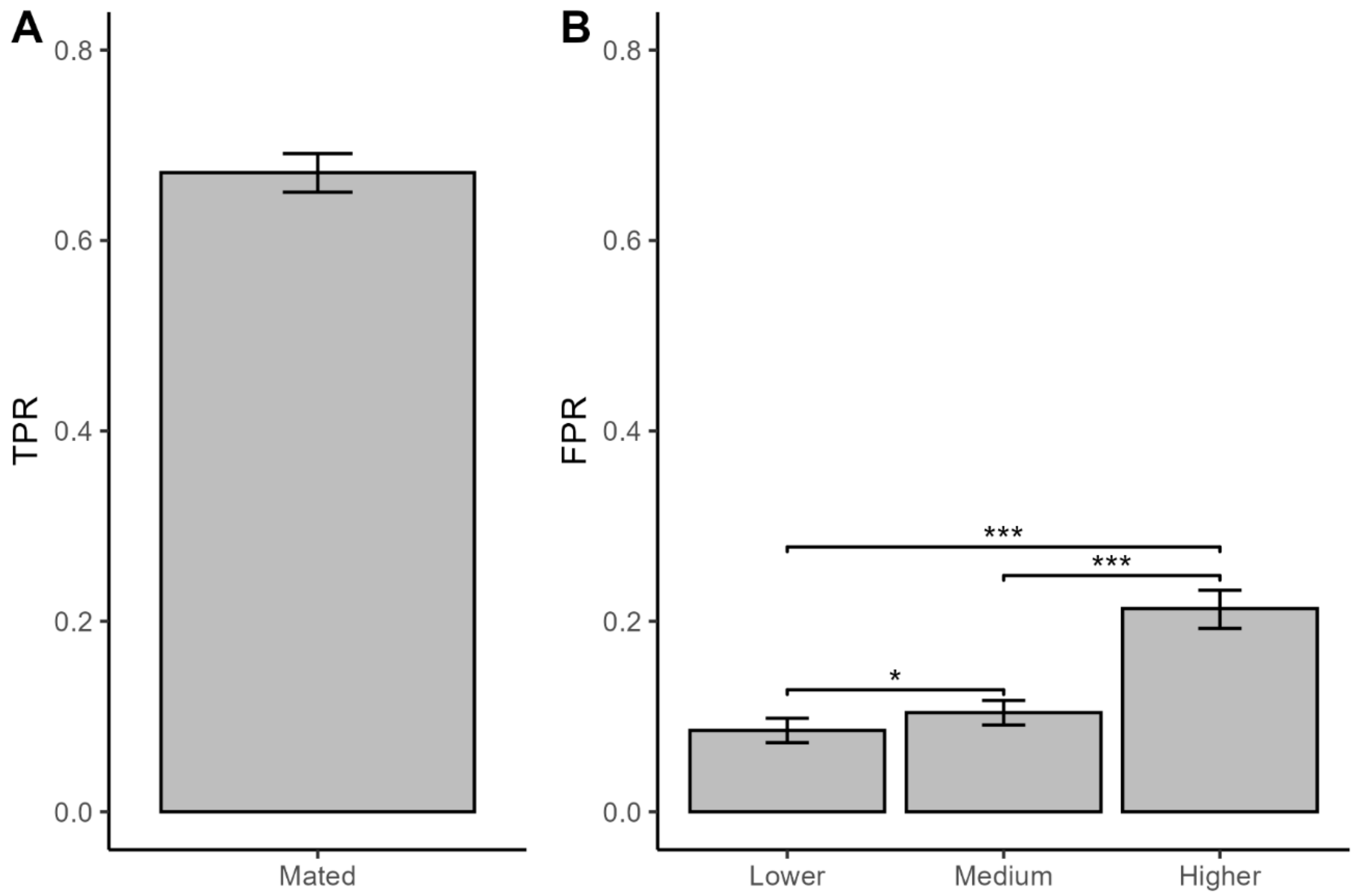

*Figure 5: (A) TPR for mated face pairs and (B) FPR for non-mated face pairs at each FPRI. Error bars are 95% confidence intervals. The TPR and FPR values are computed at the "slightly strict" reviewer decision threshold of θ = 0.5. Asterisks indicate significance: * p < 0.05, *** p < 0.001.*

Human reviewers performed worse on mated pairs (TPR = 0.67) compared to the non-mated pairs (FPRs ranging from 0.085 to 0.21), suggesting a response bias toward the "different people" decision (Figure 5). Since the mated pairs were above the 1e-6 FMR setting they all had higher similarity scores than any of the non-mated pairs. Significant differences in reviewer FPR across the FPRIs of non-mated pairs was determined through a Friedman test ($\chi^2(2) = 180.96$, $p < 0.001$). Pairwise Wilcoxon signed-rank tests with a Bonferroni correction determined significant increases in FPR when comparing the lower FPRI to the medium ($p < 0.05$) and higher ($p < 0.001$) FPRIs. There was also a significant increase from the medium to the higher FPRI ($p < 0.001$).

Statistical significance of SDT metrics was evaluated through bootstrap resampling to estimate 95% confidence intervals. Estimation of the confidence intervals was necessary due to how SDT metrics are calculated. SDT metrics for individual subjects cannot be calculated since TPR or FPR values of zero or one are not defined. Our analysis showed overlaps in the confidence intervals of sensitivity when comparing the lower and medium FPRIs, indicating that the higher similarity scores across these two intervals did not have an impact on the human reviewers' ability to correctly identify non-mated pairs (see Table 3).

*Table 3: Sensitivity and criterion for each FPRI at a "slightly strict" reviewer decision threshold (θ = 0.5)*

| FPRI | *d'* (95% CI) | *c* (95% CI) |
|---|---|---|
| Lower | 1.81 (1.72 - 1.91) | 0.46 (0.41 - 0.51) |
| Medium | 1.70 (1.62 - 1.78) | 0.41 (0.36 - 0.46) |
| Higher | 1.24 (1.16 - 1.31) | 0.18 (0.13 - 0.22) |

However, the reviewers' performance was statistically and notably worse on pairs from the higher FPRI compared to both the lower and medium FPRIs. Additionally, human reviewers used a less conservative criterion in labeling a face pair from the higher FPRI as non-mated compared to the lower and medium FPRIs (i.e., reviewer responses were less cognitively biased towards labeling a pair in the higher FPRI as "different people").

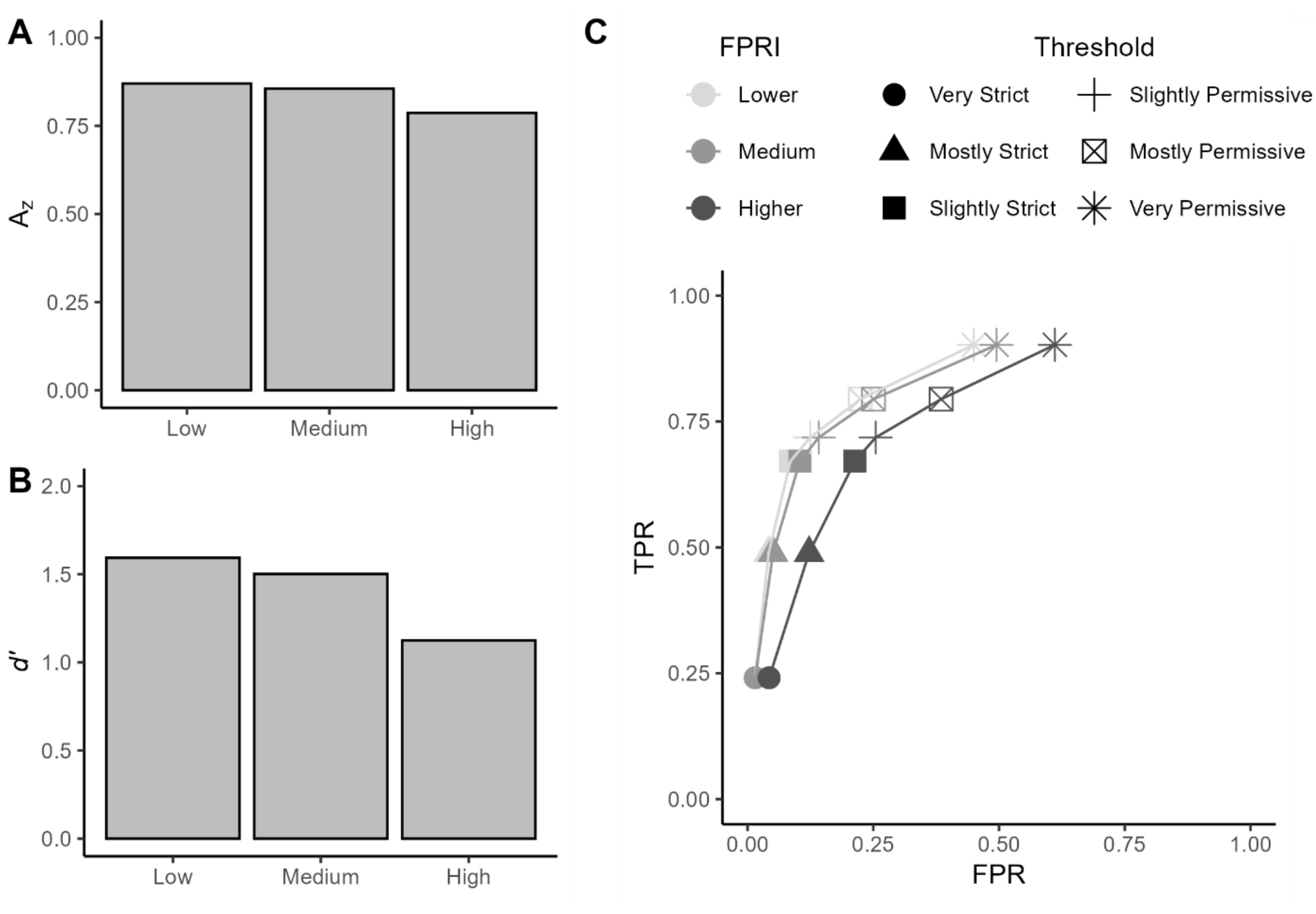

*Figure 6: (A) Area under the ROC curve ($A_z$), (B) sensitivity (d') calculated across reviewer decision thresholds, and (C) ROC curve at each FPRI.*

Examination of $A_z$ and overall sensitivity (collapsed across reviewer decision thresholds) revealed decreases in both metrics as the similarity scores between non-mated pairs increased (Figures 6A and 6B). The ROC curves shown in Figure 6C highlight the overlap in reviewer performance for the low and medium FPRIs. There is a notable shift to the right for the high FPRI ROC curve indicating an increase in the FPR at each of the evaluated reviewer decision thresholds.

**Disaggregated by Algorithm**

We examined reviewer performance for each of the four commercial algorithms separately to determine if there was an effect of algorithm. Approximately the same number of responses were included in the analysis (surveys for algorithms A, C and D each had responses from 201 reviewers and the survey for algorithm B had 198 responses). All results are reported for the slightly strict reviewer decision threshold (θ = 0.5). Figure 7A shows comparable TPR by human reviewers on the mated pairs across algorithms ($p = 0.076$); given the overlap in mated pairs for each algorithm this confirms the use of mated pairs to avoid habituation. There was a minimal increase in FPR for algorithms A and B when comparing the lower and medium FPRI; however, there were larger increases in FPR for all algorithms when comparing the medium to higher FPRI (Figure 7B).

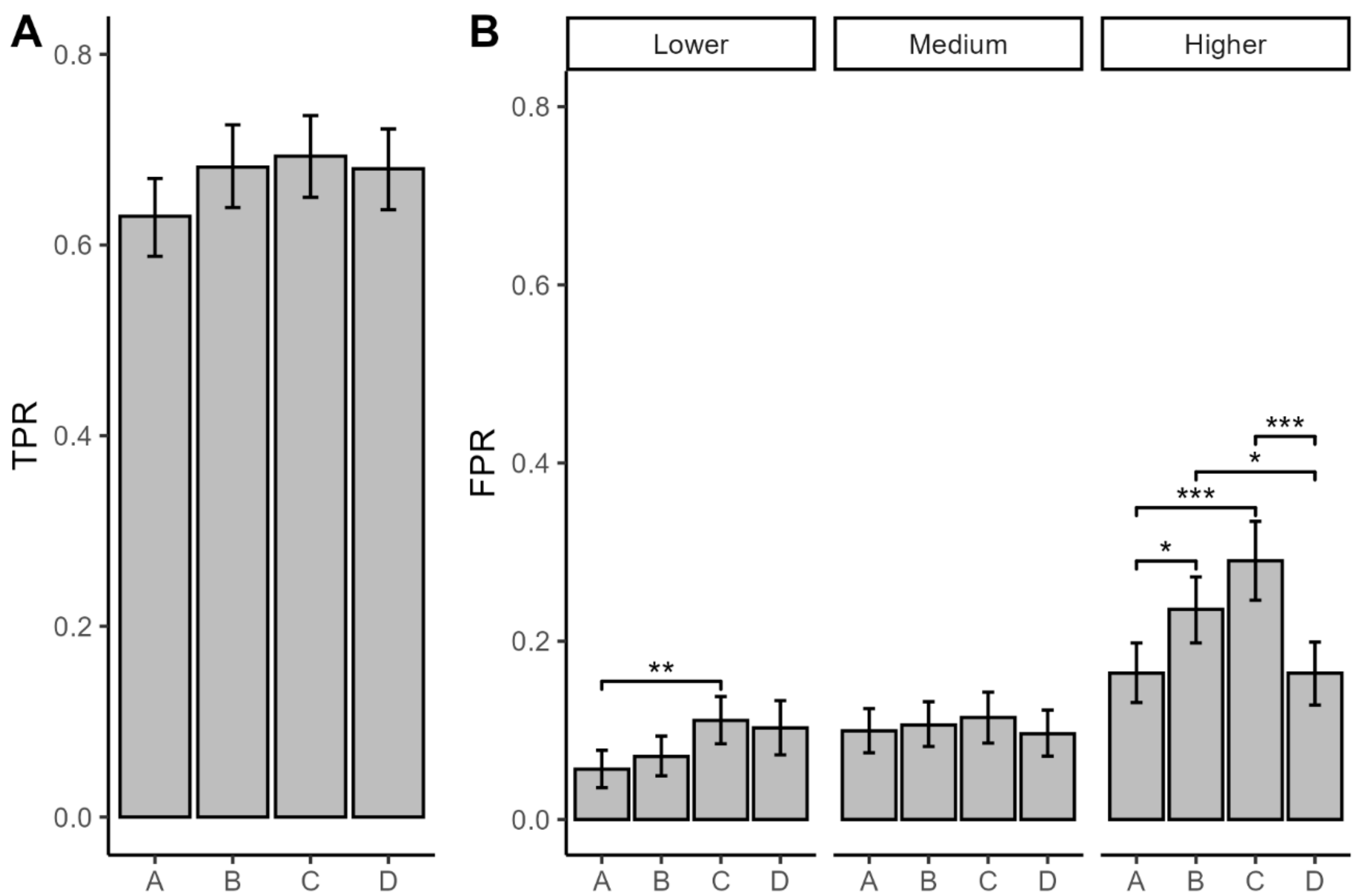

*Figure 7: (A) TPR for mated face pairs by algorithm and (B) FPR for non-mated face pairs by FPRI and algorithm. Error bars are 95% confidence intervals. The TPR and FPR values are computed at the "slightly strict" reviewer decision threshold of θ = 0.5. Asterisks indicate significance: * p < 0.05, ** p < 0.01, *** p < 0.001.*

A Kruskal-Wallis rank sum test revealed a significant main effect across algorithms within the low ($\chi^2(3) = 14.45$, $p < 0.01$) and high ($\chi^2(3) = 29.16$, $p < 0.001$) FPRIs for FPR. No significant differences were found for the medium FPRI ($p = 0.74$). A pairwise Wilcoxon signed-rank test with a Bonferroni correction applied to the low FPRI showed a significant increase in FPR between algorithm A and C ($p < 0.01$). For the high FPRI, we found significant differences between several algorithms. The FPR for algorithm A was significantly lower than for algorithms B ($p < 0.05$) and C ($p < 0.001$). The FPR for algorithm D was also significantly lower than algorithm B ($p < 0.05$) and C ($p < 0.001$).

At a slightly strict reviewer decision threshold (θ = 0.5), the SDT analysis of each algorithm revealed overlaps in the sensitivity and criterion confidence intervals for the low and medium FPRIs indicating no differences between algorithms for these intervals (Table 4). Within the high FPRI we found that the confidence intervals for algorithms C and D do not overlap for sensitivity and criterion. Overall, the sensitivity for algorithm C is lower compared to the other algorithms indicating that the perceptual similarity of the non-mated pairs selected by this algorithm was the most difficult to distinguish from the perceptual similarity of the mated pairs. The criterion for algorithm C was also lower compared to the other algorithms. This indicates that human reviewers applied a less conservative criterion to image pairs selected by algorithm C, i.e. they were less cognitively biased to respond "different people" for non-mated pairs selected by that algorithm.

*Table 4: Sensitivity and criterion for each FPRI and algorithm at "slightly strict" reviewer decision threshold (θ = 0.5)*

| FPRI | Algorithm | *d'* (95% CI) | *c* (95% CI) |
|---|---|---|---|
| Lower | A | 1.92 (1.69 - 2.13) | 0.63 (0.51 - 0.73) |
| | B | 1.94 (1.74 - 2.13) | 0.50 (0.38 - 0.60) |
| | C | 1.73 (1.55 - 1.90) | 0.36 (0.26 - 0.46) |
| | D | 1.73 (1.53 - 1.93) | 0.40 (0.29 - 0.51) |
| Medium | A | 1.62 (1.45 - 1.78) | 0.48 (0.38 - 0.57) |
| | B | 1.72 (1.53 - 1.89) | 0.39 (0.29 - 0.48) |
| | C | 1.71 (1.52 - 1.88) | 0.35 (0.25 - 0.45) |
| | D | 1.77 (1.60 - 1.93) | 0.42 (0.31 - 0.52) |
| Higher | A | 1.31 (1.15 - 1.46) | 0.32 (0.22 - 0.42) |
| | B | 1.19 (1.06 - 1.32) | 0.12 (0.03 - 0.22) |
| | C | 1.06 (0.90 - 1.21) | 0.02 (-0.08 - 0.12) |
| | D | 1.44 (1.27 - 1.60) | 0.25 (0.15 - 0.36) |

Across algorithms there is a decrease in $A_z$ and sensitivity from the lower to higher FPRIs indicating a decrease in performance by the reviewers suggesting a decreasing ability of human reviewers to detect algorithm errors at higher FMR settings (Figure 8). When comparing the lower and medium FPRIs, decreases in $A_z$ and *d'* are observed for algorithm A, while these metrics are relatively stable for the other algorithms. The largest decrease for $A_z$ and *d'* is seen in algorithm C between the medium and high FPRIs. The human ROC curve for each algorithm visualizes these findings (Figure 8C). For algorithms B, C, and D there is an overlap of the ROC in the lower and medium FPRIs. Notably, the human ROC curve for the higher FPRI with algorithm C drastically shifts to the right in comparison to the other algorithms, showing the observed increases in the FPR.

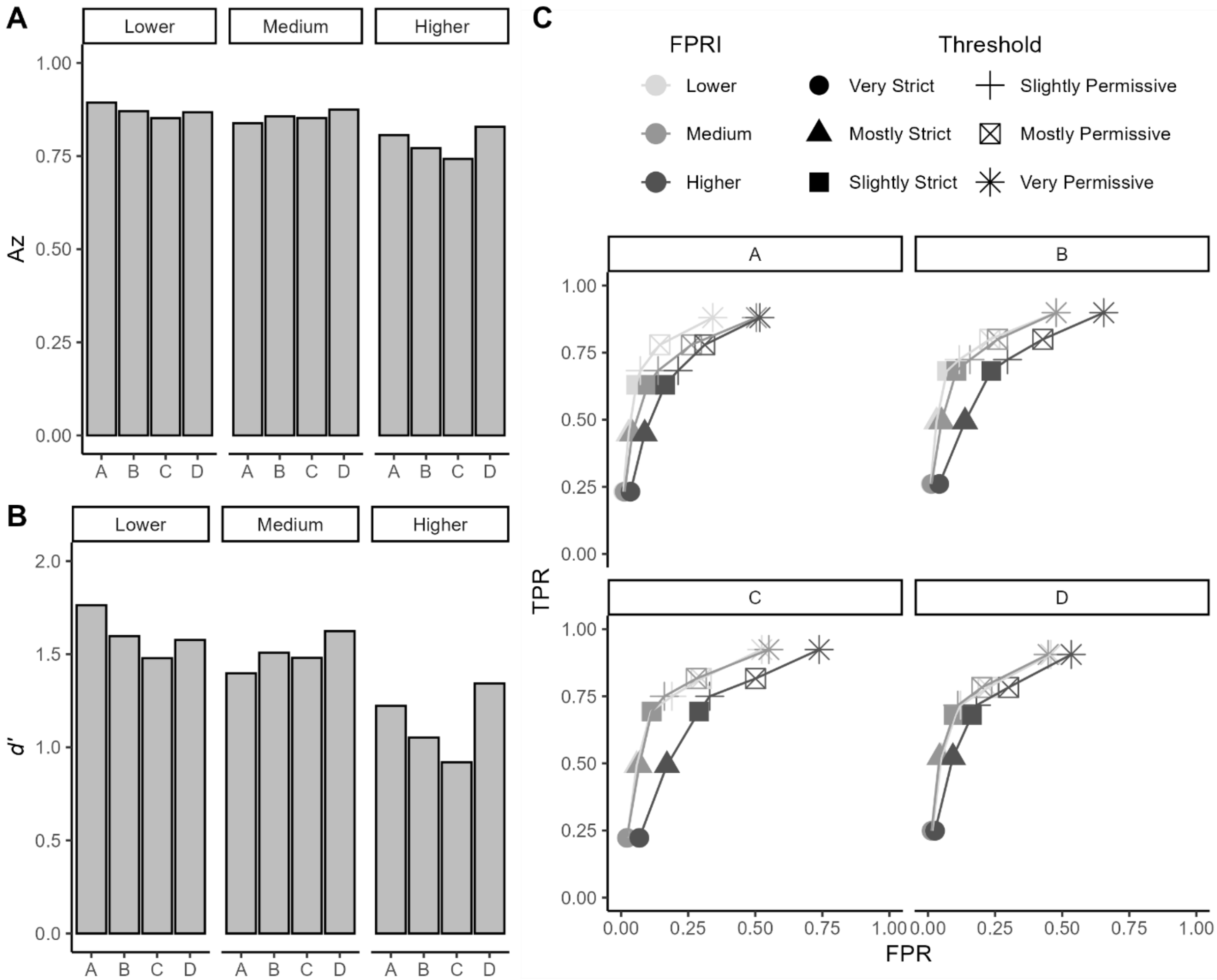

*Figure 8: (A) Area under the ROC curve ($A_z$), (B) sensitivity (d') calculated across reviewer decision thresholds, and (C) ROC curve at each FPRI for each algorithm.*

## DISCUSSION

Increased deployment of FR algorithms in identity workflows offers significant benefits in terms of scale, accuracy, and automation. However, as algorithmic performance improves, human reviewers face greater challenges in providing effective oversight, particularly in high-stakes, high-security settings such as law enforcement. This study addresses the growing need to measure human capabilities in AI workflows to design better human-AI teams and to manage the risks posed by AI systems.

We found that human review can significantly reduce false positive errors made by commercial face recognition algorithms under a range of settings. At the highest security setting associated with the lowest FMR, untrained human reviewers detected ~80% of algorithmic false positives with moderate sensitivity (d' ~ 1). It is likely that trained examiners would have an even greater success rate [39,56]. Our findings expand on those from Towler and colleagues' (2023) that human experts and FR algorithms can use different features to discriminate face photos. The fact that even untrained human reviewers can detect algorithmic false positives reinforces the value of human review in face recognition technology workflows as a critical component of oversight and risk management for these AI systems.

At higher security settings, algorithms are configured to prioritize minimizing false positives. We found that stricter security configurations reduce the number of algorithmic errors yet amplify the difficulty for human reviewers to detect algorithmic false positives. Our findings highlight the unintended consequence of stricter security settings in workflows that rely on human oversight, increasing risks for high-security applications such as law enforcement, where a single false positive can have severe consequences. We showed that human reviewers are progressively less effective at catching FR algorithm false positive errors as algorithms are configured to operate at higher security settings. When operating at the highest biometric threshold used for non-mated pairs (FMR = 1e-5), human reviewers' false positive detection rates dropped by more than half compared to lower thresholds. This suggests that system designers must carefully consider how such security configurations impact human oversight, including training requirements, during algorithm selection and workflow design.

Human reviewers' ability to detect false positives varied significantly between algorithms even when the algorithms had comparable performance (FMR and false non-match rate (FNMR)). This variability underscores the need to evaluate not just algorithmic accuracy, but also how compatible an algorithm is with human review during operational deployment. Reviewer FPRs varied nearly twofold across the four commercial algorithms examined. This shows that measures of human-algorithm compatibility should be considered when selecting algorithms for deployment, especially for workflows involving manual review. This represents a shift in evaluation metrics, moving beyond algorithmic benchmarks alone to include human-AI teamwork metrics.

Existing biometric testing standards, such as the ISO/IEC 19795 series, focus on algorithmic accuracy. While some parts of ISO/IEC 19795 address the role of human "attendants" or operational personnel (e.g., ISO/IEC 19795-6 [57]), these standards do not address the human review in biometric systems. Human review is addressed in forensic guidelines from the Facial Identification Scientific Working Group (FISWG) which define the roles of a "facial reviewer" of algorithmic candidate lists [58], however, these guidelines consider review being independent of the algorithm and focus on procedures rather than total system performance. Given the demonstrated sequential interdependence between algorithms and human reviewers, incorporating notions of human-AI team performance into evaluation frameworks [46] would help organizations develop a more comprehensive understanding of the overall effectiveness of these systems.

Researchers have long compared human and algorithm performance, including measuring human performance using face recognition datasets on which performance of some algorithms was low [15,59–61]. Early face recognition algorithms were challenged by degradations in image quality due to pose, expression, and illumination. These perturbations primarily affected mated similarity scores resulting in false negative errors. For example, the Good, the Bad, the Ugly (GBU) dataset was based on the ranking of mated similarity scores with the goal of looking at the effects of image quality to compare human and algorithm performance [59]. However, variations in face image quality likely have a distinct influence on human perception relative to variations in face feature similarity. Our work used images which were of uniformly high quality (frontal pose, neutral expression, diffuse illumination, neutral background). We developed and used the S-FPRI method to identify image pairs that varied based on the non-mated similarity scores in a way that that is both operationally relevant and facilitates comparisons across current commercial algorithms. Our focus in this work is not to compare human and algorithm performance, rather we examine the ability of humans to detect algorithmic false positives on the basis of the degree of similarity of the innate biometric features of the biometric capture subjects as assessed by current commercial biometric algorithms. Understanding the ability of human reviewers to detect specific algorithmic errors is an important factor in developing successful human-AI teams.

Interestingly, the difficulty of the face pairs selected for the DFICT dataset used in this study was similar or greater than the difficulty measured using a human driven down-selection process. For example, when aggregating human performance across the four algorithms, sensitivity measured here was comparable to the short forms of the updated Glasgow Face Matching Test (GFMT2) [62]. White & colleagues (2022) used an open-source FR algorithm (VGGFace2), followed by expert human review and novice human ratings to create the GFMT2, which they found to have a *d'* range of 1.64-1.75 across two 40-item short form versions and a combined 80-item version. This range is within the *d'* range we found across the FPRIs defined by algorithmic FMR settings (1.24 to 1.81). Comparing GFMT2 sensitivity to the high FPRI (*d'* = 1.24) suggests that face pairs selected by algorithms at FMR settings of 1e-5 or lower are more challenging for human reviewers compared to face pairs selected by other humans.

Prior work has demonstrated that new face recognition algorithms can increase the difficulty of human review [45]. Our results show that the level of difficulty introduced can vary depending on the specific algorithm used as well as based on the configured security settings. Newer algorithms can be configured with biometric thresholds that allow them to produce acceptable recognition performance at lower false match rates. Our work shows that these higher security settings can increase the false positive rates in human review. False positive rates for human reviewers more than doubled from under 10% at the lower security FPRI (FMR setting between 1e-3 and 1e-4) to over 20% at the higher security FPRI (FMR setting between 1e-5 and 1e-6). Thus, when biometric algorithms are configured for facilitation, such as in identity verification use cases in the travel environment, human reviewers may be capable of detecting algorithmic false positives even without formal training on face recognition tasks. On the other hand, when biometric algorithms are configured for greater security, such as in law enforcement identifications, human reviewers without formal training would be less likely to reject false positives [45,63]. This shows the importance of considering both the algorithm and its configuration as well as the level of reviewer expertise required when designing biometric use cases involving human review.

In our experimental design, we did not present human reviewers with any information about the algorithm's accuracy or its decisions about each face pair. This allowed us to assess human reviewer performance without introducing response bias from the algorithm's decisions [40–44]. However, in operational settings human reviewers are often provided with information from the algorithm, which can potentially reduce the likelihood of the reviewer identifying algorithmic false positives. Prior research has shown that when working with decision-aid systems, the type of information and how much information is presented can influence human decisions [64–66]. The current advancements of AI, specifically large language models (LLMs), have created the possibility for high levels of communication and coordination between a human and AI teammate beyond simple binary outcomes or scores, which could greatly improve information sharing between teammates. While prior work can somewhat inform the level of transparency required for a human-AI team utilizing LLMs, additional research needs to focus on how to leverage conversational AI to improve team performance.

Our study comes with important limitations. All images used in this study were collected under conditions that controlled for variation in performance due to differences in biometric sample quality. The quality of images in operational settings is likely worse than those used in this study [29]. Both human and algorithm face recognition performance is known to be affected by the quality of the utilized biometric samples and future work should investigate whether similar results hold for images of operational quality. The task we used is reflective of a verification use case, not identification use cases that require review of a candidate list [45]. We expect that the reduction in human reviewer performance we observed for verification would extend to identification as well. Our study was limited by the size of the reference dataset used to identify similar face pairs, which is small in comparison to the galleries used in operational settings. Larger galleries afford more non-mated pairs, leading to an increased likelihood of finding extremely similar individuals. Our data was limited in finding false positives at FMR settings below 1e-6 and these pairs were not included in our survey. We expect that human reviewer performance at these higher security settings would be even lower. Finally, due to logistical limitations, our study was limited in the number of face pairs that could be shown to individual human reviewers. Thus, the ratio of nine non-mated face pairs and three mated face pairs per reviewer may have introduced a response bias in favor of a "different person" response which is specific to this experimental design and would not necessarily reflect a specific applied use case. Future work should further examine the suitability of specific algorithms for human-AI teams using a greater sample of face pairs in specific FPRIs.

Our work carries important implications for human algorithm teaming in applied use-cases. We have shown that the selected algorithm and security configurations can influence the performance of human reviewers. When developing a human-AI team within the identity workflow, teaming heuristics need to account for these factors in addition to the specific use case and required level of human expertise. Furthermore, our results show the importance of evaluating human reviewers using datasets that represent algorithmic false positives in a calibrated manner. To date, biometric algorithms have been largely selected based on their technical accuracy benchmarks alone and biometric testing standards do not consider combined human and algorithm performance. Our work suggests that similarly accurate algorithms can nonetheless have major differences in their compatibility with human review, perhaps due to differences in the overlap between the biometric features utilized by human reviewers and these AI systems. We suggest that, in applications where human review is necessary, the ability of human reviewers to catch algorithmic errors should be considered as a factor in algorithm selection.

## ACKNOWLEDGMENTS

This research was sponsored by the United State Department of Homeland Security, Science and Technology Directorate on contract numbers 70RSAT23CB0000003. CRediT author contributions: EE: methodology, investigation, writing – review and editing; JJH: conceptualization, methodology, writing – review and editing; IMS: conceptualization, project administration, methodology, investigation, data curation, formal analysis, visualization, writing – original draft; YBS: conceptualization, supervision, methodology, visualization, writing – original draft; RW methodology, data curation, formal analysis, writing – review and editing.